\documentclass[10pt,twocolumn,letterpaper]{article}

\usepackage{multirow}
\usepackage{pifont}
\usepackage{xcolor}
\usepackage{tabularx}
\usepackage{algorithm}
\usepackage{algpseudocode}
\usepackage[normalem]{ulem}

\usepackage{adjustbox}
\usepackage[pagenumbers]{wacv}              
\usepackage[accsupp]{axessibility}  

\newcommand{\TRCQA}{TRIFLE}
\newcommand{\TR}{PRISM}

\definecolor{wacvblue}{rgb}{0.21,0.49,0.74}
\usepackage[pagebackref,breaklinks,colorlinks,allcolors=wacvblue]{hyperref}

\def\wacvPaperID{1053} 
\def\confName{WACV}
\def\confYear{2027}

\title{\TR: Pan-Script Table Structure Recognition by Invariant Spatial Modeling}

\author{
Jahanvi Rajput$^{1}$,
Dhruv Kudale$^{1}$\thanks{Work done while pursuing MS at IIT Bombay.}, 
Nitin Singh Patel$^{1}$,
Saikiran Kasturi$^{1}$,
Ganesh Ramakrishnan$^{1,2}$ \\
{\tt\small \{23d0378, 22m2116, 22b0974, 24m2157, ganramkr\}@iitb.ac.in} \\[4pt]
$^{1}$Indian Institute of Technology Bombay, $^{2}$BharatGen}

\begin{document}
\maketitle
\begin{abstract}
Table Structure Recognition (TSR) is a fundamental task in document intelligence, underpinning downstream applications such as Table Reconstruction (TR) and Complex Table Reasoning (CTR). Existing TSR methods are largely English-centric and often rely on script-specific textual cues, limiting their generalization to non-Latin languages. We present PRISM, a decoupled multilingual TSR pipeline that integrates physical and logical structures to generate faithful table reconstructions, enabling robust multilingual TR and CTR. At its core, PRISM employs SPECTRUM, a script-agnostic Physical TSR model. SPECTRUM learns geometry-centric representations through masking-based augmentation and substantially improves cross-script generalization without requiring multilingual supervision. We demonstrate that the improved TSR achieved by PRISM consistently translates to downstream gains in both TR and CTR. To facilitate comprehensive evaluation, we introduce two first-of-their-kind multilingual benchmarks: CUSTARD (1,215 TSR-annotated tables in 12 non-Latin languages) and TRIFLE (364 diverse non-Latin tables for TR and CTR). Extensive experiments show that PRISM improves multilingual TSR by 4.9 p.p., yielding downstream gains of 3.5 p.p. in TR and 2.8 p.p. in CTR while maintaining competitive performance on standard English benchmarks.  We release our code, models, and benchmark at: https://github.com/JahanviRajputTIH/PRISM.git.

\end{abstract}
    
\section{Introduction}
\label{sec:intro}

\begin{figure*}[t]
    \centering
    \includegraphics[width=0.85\textwidth]{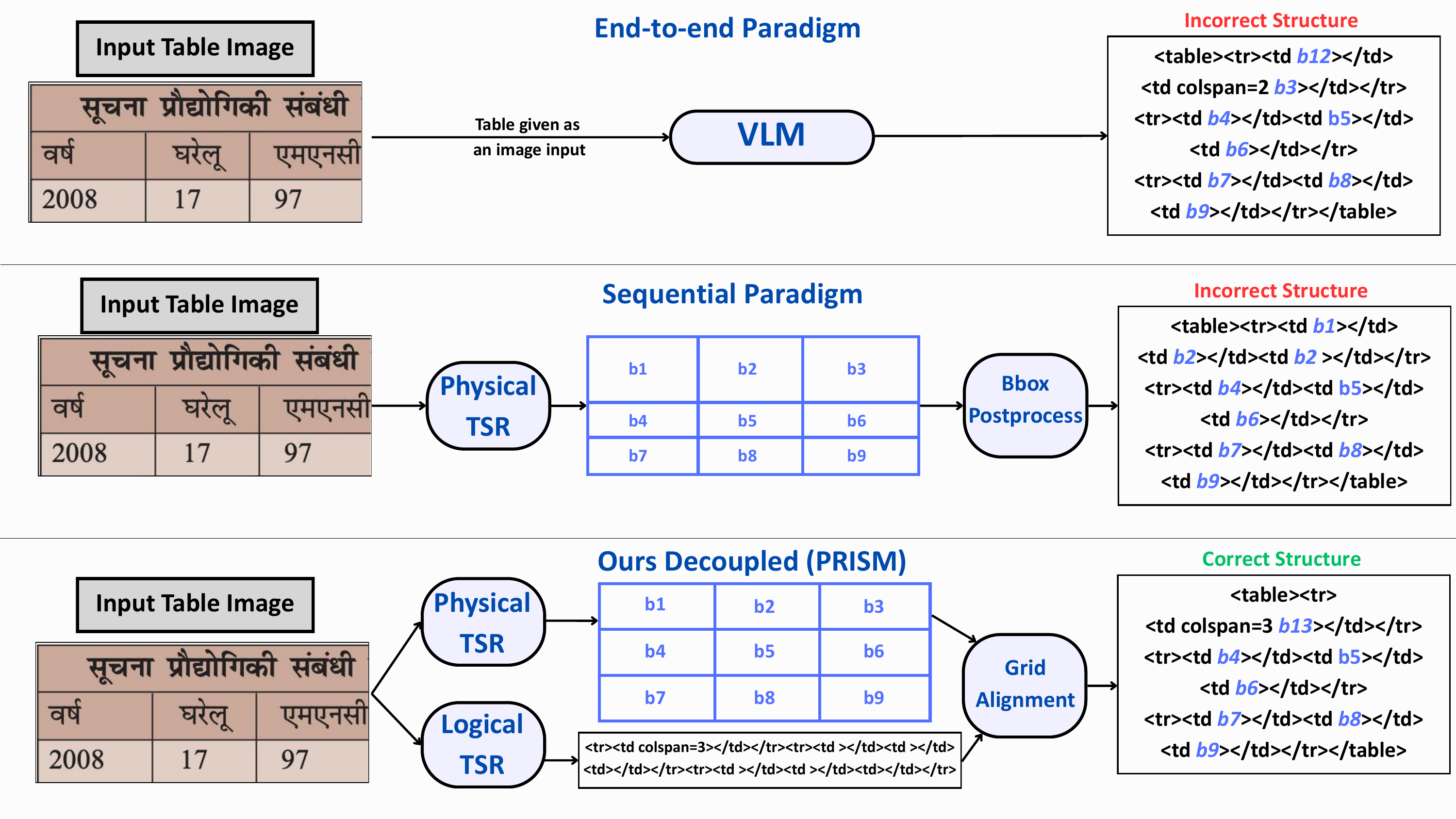}
    \caption{End-to-end TSR entangles text and structure, limiting multilingual generalization, while sequential methods rely on fragile post-processing and poorly handle spanning cells. Our decoupled paradigm separates physical and logical TSR to achieve script-agnostic TSR.
    }
    \label{fig:spectrum_teaser_fig}
\end{figure*}

\begin{figure}[!htb]
    \centering  
    \includegraphics[width=0.48\textwidth]{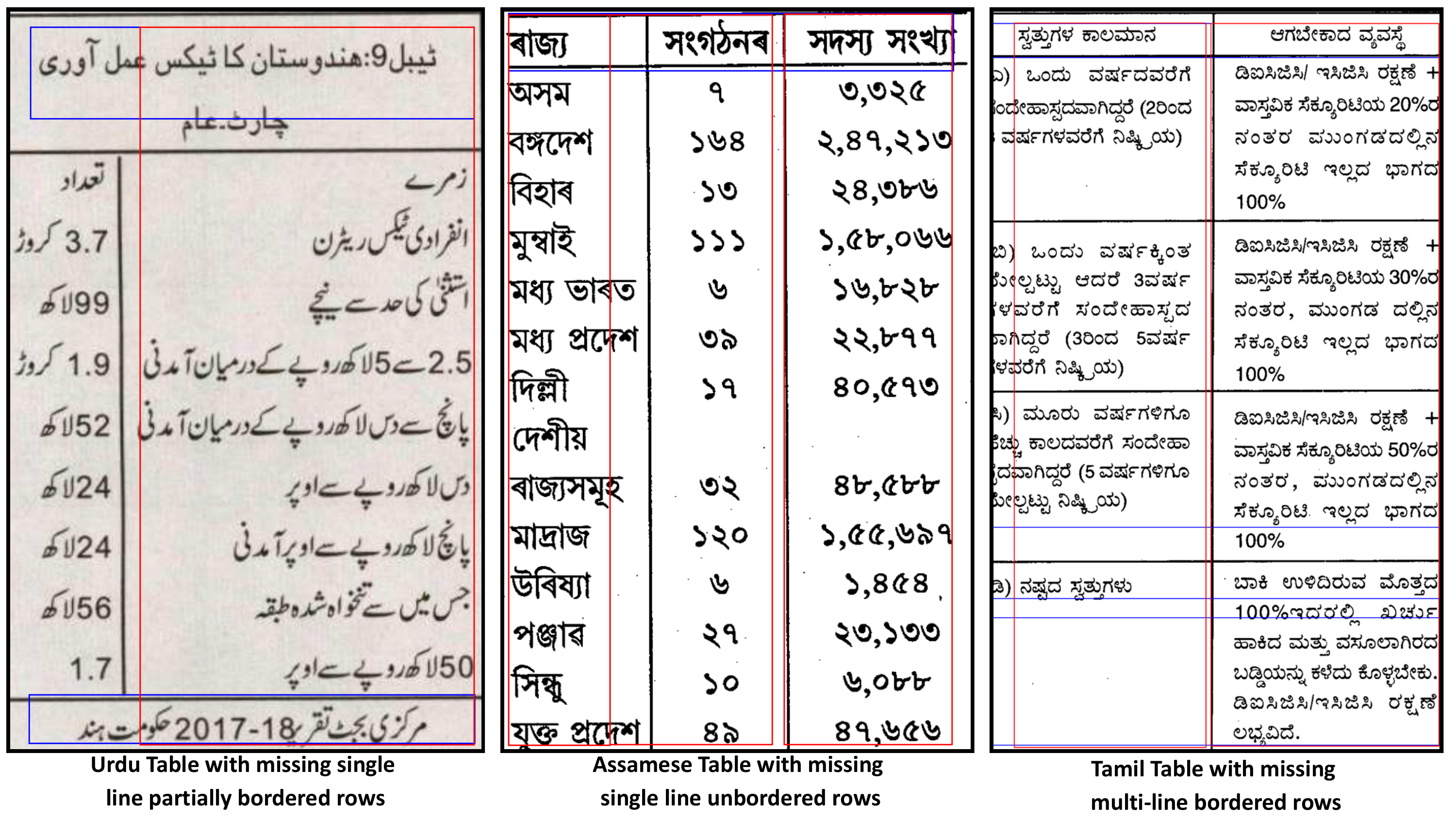}
    \caption{Examples using TATR \cite{tatr-pub-1m} showing that rows (blue) are often missed in partially bordered or unbordered tables, or tables with multi-line rows, showing its sensitivity to script variations.}
    \label{fig:bad_examples}
\end{figure}

Table Structure Recognition (TSR) is a fundamental task in document intelligence that aims to recover the structural organization of tables from document images by identifying their constituent rows, columns, cell boundaries, and relationships. Accurate TSR forms the foundation for downstream tasks such as Table Reconstruction (TR), where table structure is combined with OCR-extracted text to reconstruct complete machine-readable tables, and Complex Table Reasoning (CTR), where answering questions often requires multi-step reasoning over both the table structure and cell contents, rather than simple value extraction. Since TSR is the foundation of the table understanding pipeline, errors introduced at this stage cascade through subsequent stages, making robust TSR critical for reliable end-to-end table understanding. Despite recent advances, existing TSR methods remain largely English-centric and generalize poorly to tables having content in non-Latin scripts. As illustrated in Figure~\ref{fig:spectrum_teaser_fig}, end-to-end TSR methods entangle text and structure, while sequential pipelines depend on brittle post-processing. Consequently, even state-of-the-art models such as TATR \cite{tatr-pub-1m} degrade substantially on multilingual tables, particularly in row detection (Figure~\ref{fig:bad_examples}). These limitations motivate the need for script-agnostic TSR frameworks.

To overcome these limitations, we propose \TR{} (\textit{Pan-Script Table Structure Recognition by Invariant Spatial Modeling}), a decoupled multilingual TSR framework that explicitly separates physical or spatial TSR (localizing rows and columns) from logical TSR (recovering cell relationships and spanning information) while maintaining structural consistency. At its core, \TR{} incorporates \name{} (\textit{Script-agnostic Extraction of Table Structure for Multilingual Tables}), a script-agnostic physical TSR model that employs masking-based augmentation to suppress script-specific textual cues and learn from alignment, spacing, and ruling lines. Specifically, \name{} predicts row and column bounding boxes, while an existing state-of-the-art script-agnostic logical TSR module independently infers cell relationships. These complementary representations are unified through grid-based alignment to generate structurally faithful HTML with grounded cell locations (Figure~\ref{fig:spectrum_teaser_fig}). These locations can be subsequently used for OCR-based cell content extraction to perform complete TR. We further demonstrate that using \TR{} consistently improves downstream tasks. To facilitate future research, we introduce two complementary benchmarks. Our contributions are as follows:

\begin{enumerate}

\item We propose \textbf{\name{}}, a script-agnostic physical TSR model that detects rows and columns as bounding boxes, enabling cross-script generalization.

\item We propose \textbf{\TR{}}, a decoupled multilingual TSR framework that unifies independently predicted physical and logical structures, and demonstrate that it consistently improves downstream TR and CTR tasks.

\item We introduce two multilingual table-based benchmarks: \textbf{\dataset{}} (\textit{Content and Unified Structures of Tables in Regional Documents}), containing 1,215 annotated tables across 12 non-Latin languages for TSR, and \textbf{\TRCQA{}} (\textit{Table Reconstruction and inFerencing for compLex quEstions}), supporting end-to-end evaluation of TR for 364 tables and CTR using 714 QA pairs on those tables in three non-Latin languages.

\end{enumerate}

\section{Related Work}
\label{sec:Related work}

\textbf{End-to-End TSR} methods employ vision-language models (VLMs) and multi-modal architectures to directly generate structured representations, such as HTML, from table images. Representative approaches include MTLTabNet \cite{mtl-tab-net}, SmolDocLing \cite{nassar2025smoldocling}, SmolVLM \cite{smolvlm}, and Granite-Vision \cite{granitevision}. By jointly modeling structure and content without explicit OCR, these methods offer a unified framework for table understanding. However, they often entangle textual appearance with structural reasoning, leading to script-dependent representations that generalize poorly to multilingual and noisy document images. \textbf{Sequential TSR} methods explicitly perform physical TSR to demarcate cells, followed by OCR for content extraction. They localize structural elements such as rows, columns, or cells using object detectors and transformer-based architectures, including Faster R-CNN \cite{faster-rcnn}, Mask R-CNN \cite{mask-rcnn}, YOLO \cite{yolo}, DETR \cite{detr}, TATR \cite{tatr-pub-1m}, TableFormer \cite{tableformer}, and TSRFormer \cite{tsrformer3}. While effective for localization, these approaches typically depend on heuristic post-processing, making them susceptible to detection errors and challenging layouts involving spanning cells. To compensate for this, image-to-sequence methods for logical structure prediction have been proposed. They directly generate HTML or LaTeX representations, including OTSL-based approaches. Examples include SPRINT \cite{otsl,kudale2025sprint}, graph-based models such as GTE \cite{zheng2021global} and TGRNet \cite{tgrnet}, and hybrid frameworks \cite{zheng2021global,tsr-local-attention,lgpma,tabstruct,vast}. 

Existing TSR datasets, including PubTabNet \cite{teds}, FinTabNet \cite{zheng2021global}, PubTables-1M \cite{tatr-pub-1m}, SciTSR \cite{scitsr}, SynthTabNet \cite{tableformer}, and TableBank \cite{tablebank}, have significantly advanced TSR through rich physical and logical annotations and standardized evaluation metrics such as TEDS \cite{teds} and GriTS \cite{grits}. However, these datasets are predominantly English-centric and do not systematically assess robustness on non-Latin scripts, motivating the introduction of our \dataset{} benchmark. \textbf{Decoupled TSR} is one of the recent advances, which include approaches like DELTA \cite{rajput2026delta}, and have established this paradigm by independently modeling physical and logical structures. However, while logical TSR is inherently script-agnostic \cite{kudale2025sprint}, the physical TSR component remains script-dependent, limiting generalization to non-Latin documents. We bridge this gap with \TR{}, a multilingual decoupled TSR framework powered by \name{}, our script-agnostic physical TSR model.

\textbf{Table Reasoning} has progressed from conventional extractive Table Visual Question Answering (TabVQA), which primarily focuses on retrieving information from table images using textual content and visual layout, to the more challenging setting of CTR, where answering a question may require multi-hop reasoning, aggregation, comparison, arithmetic operations, and structural understanding of the table. Recent TabVQA approaches such as MMTabQA \cite{mmtabqa2024}, MMSci \cite{yang2025table_source_matter}, and TabPedia \cite{zhao2024tabpedia} leverage multi-modal reasoning and VLMs, achieving strong performance on extractive question answering. However, relatively little attention has been devoted to complex reasoning over tables, particularly in multilingual settings. To evaluate deeper reasoning capabilities, benchmarks such as TAT-DQA \cite{zhu2022tatdqa} and CRT-QA \cite{zhang2023crtqa}, along with methods including Chain-of-Table \cite{wang2024chainoftable}, TaPERA \cite{zhao2024tapera}, MMTabQA \cite{mmtabqa2024}, PrivTab \cite{privtab2025}, MultiTabQA \cite{multitabqa2023}, and TQA-Bench \cite{tqabench2024}, have explored increasingly sophisticated reasoning strategies over structured tables. Nevertheless, existing benchmarks remain limited in either reasoning complexity or linguistic diversity. For example, TabPedia \cite{zhao2024tabpedia} introduces multi-step questions involving arithmetic, aggregation, and logical inference, but is restricted to English tables. Conversely, TORQUE \cite{rajput2026delta} extends TableVQA to multilingual and non-English settings, yet its questions are predominantly extractive and do not adequately evaluate multi-hop complex reasoning. This leaves a significant gap at the intersection of multilingual table understanding and complex reasoning. To bridge this gap, we introduce \TRCQA{}, a benchmark for CTR over non-Latin multilingual tables, designed to evaluate both structural understanding and multi-step reasoning across diverse languages and document types.

\section{\name{} and \TR{}}
\label{sec:methodology}
We present \name{} for physical TSR and show how it is incorporated in \TR{}, our proposed TSR framework.

\begin{figure*}[!htb]
    \centering
    \includegraphics[width=0.85\textwidth]{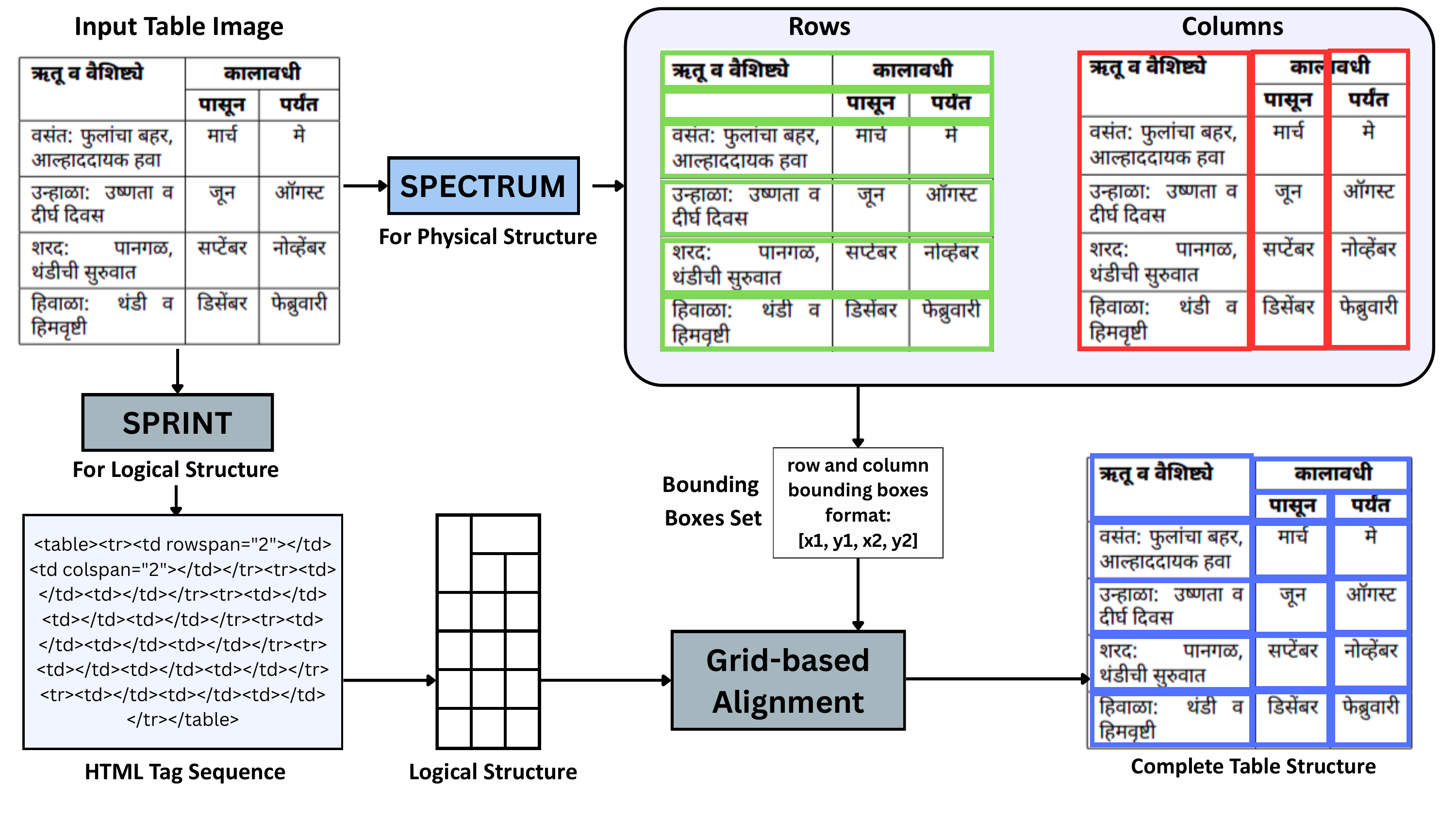}
    \caption{
    Overview of \TR{}, our TSR framework. \name{} predicts the row and column bounding boxes; SPRINT predicts the HTML corresponding to the logical structure. Grid-based alignment is performed to retrieve the complete, consistent structure of the table.}
    \label{fig:spectrum_methodology}
\end{figure*}

\subsection{\textbf{\name}}
\label{sec:spectrum}
\name{} is our proposed script-agnostic Physical TSR framework, motivated by the observation that although scripts differ in glyphs, writing styles, and typography, the geometric organization of tables remains largely invariant. Regardless of language, tables are defined by structural cues such as row and column alignment, cell spacing, and ruling lines. However, existing TSR models often exploit textual appearance as a shortcut for structure prediction, limiting their generalization to non-Latin scripts. To overcome this, \name{} fine-tunes an existing physical TSR model using masking-based augmentation, where textual regions are selectively masked while preserving the table layout. This encourages the model to learn geometry-centric rather than script-dependent representations, enabling robust localization of rows and columns without architectural modifications. It also eliminates multilingual pretraining or any script-specific supervision. As the physical TSR module of \TR{}, \name{} provides the foundation for robust multilingual TSR and its downstream applications.

\subsection{\TR{}}
\label{sec:prism}
\TR{} is our proposed decoupled framework for multilingual TSR. Its key insight is a divide-and-conquer strategy: physical and logical table structures are modeled independently before being unified into a complete table structure representation. The physical branch employs \name{}, while the logical branch leverages SPRINT~\cite{kudale2025sprint}. Since both components are inherently script-agnostic, \TR{} naturally generalizes across diverse writing systems. As illustrated in Figure~\ref{fig:spectrum_methodology}, \TR{} comprises three stages:

\begin{itemize}

\item \textbf{Physical TSR using \name{}:} \name{} recovers the table geometry by detecting row and column bounding boxes. Through masking-based fine-tuning, it learns geometry-centric representations, enabling robust localization across scripts without multilingual pretraining or script-specific supervision.

\item \textbf{Logical TSR using SPRINT:} SPRINT~\cite{kudale2025sprint} predicts the logical table structure, including cell ordering, spanning patterns, and row-column relationships, complementing the geometric representation recovered by \name{}.

\item \textbf{Grid-based Alignment:} The physical and logical structures are reconciled through grid-based alignment as proposed in SPRINT \cite{kudale2025sprint}, where logical cells are mapped onto the geometric grid while prioritizing the physical structure whenever conflicts arise. The aligned representation is converted into attributed HTML, with each \texttt{<td>} associated with its grounded cell bounding box.

\end{itemize}

For TR, the complete TSR representation produced by \TR{} is combined with cell-wise OCR to populate each recovered cell, resulting in a complete machine-readable HTML table. For CTR, this reconstructed HTML serves as structured input to downstream reasoning models, enabling multi-step reasoning over both table topology and cell contents rather than simple value extraction.

\section{\dataset{} and \TRCQA{}}
We introduce \dataset{} and \TRCQA{}, two multilingual benchmarks for evaluating TSR and its downstream tasks.

\subsection{\textbf{\dataset{}}}
We introduce \dataset{}, a manually curated benchmark for multilingual TSR comprising 1,215 tables across 12 non-Latin languages. Built upon MUSTARD~\cite{kudale2025sprint}, which provides logical structure annotations in OTSL format, \dataset{} augments every table with manually annotated row and column bounding boxes, enabling comprehensive evaluation of both physical and logical TSR. The language-wise distribution is summarized in Table~\ref{tab:dataset_stats}. \dataset{} captures substantial linguistic and structural diversity, including bordered, partially bordered, and borderless tables, spanning cells, multi-line entries, and irregular layouts commonly encountered in real-world documents (Figure~\ref{fig:dataset_samples}). All annotations were performed manually by six annotators and independently verified by two individuals following a standardized protocol. For bordered tables, cell boundaries were aligned with the ruling lines, whereas for partially bordered and borderless tables, the underlying grid was inferred from cell alignment, spacing, and layout regularity. This protocol ensured consistent structural annotations. 


\begin{table}[!htb]
\centering
\setlength{\tabcolsep}{8pt}
\renewcommand{\arraystretch}{1.1}
\begin{tabular}{cccc}
\hline
\textbf{Language} & \textbf{Count} & 
\textbf{Language} & \textbf{Count} \\ 
\hline
Assamese  & 101 & Hindi     & 101  \\
Bengali   & 100 & Kannada   & 101  \\
Chinese   & 102 & Malayalam & 102  \\
Gujarati  & 101 & Oriya     & 101  \\
Punjabi   & 102 & Urdu      & 102 \\
Tamil     & 100 & Telugu    & 102 \\
\hline
\end{tabular}
\caption{\dataset{} language-wise distribution of 1,215 tables.}
\label{tab:dataset_stats}
\end{table}


\subsection{\TRCQA{}}
\begin{figure}[t]
    \centering

    \begin{subfigure}{\linewidth}
        \centering
        \includegraphics[width=0.85\linewidth,trim={0 0 13cm 0},clip]{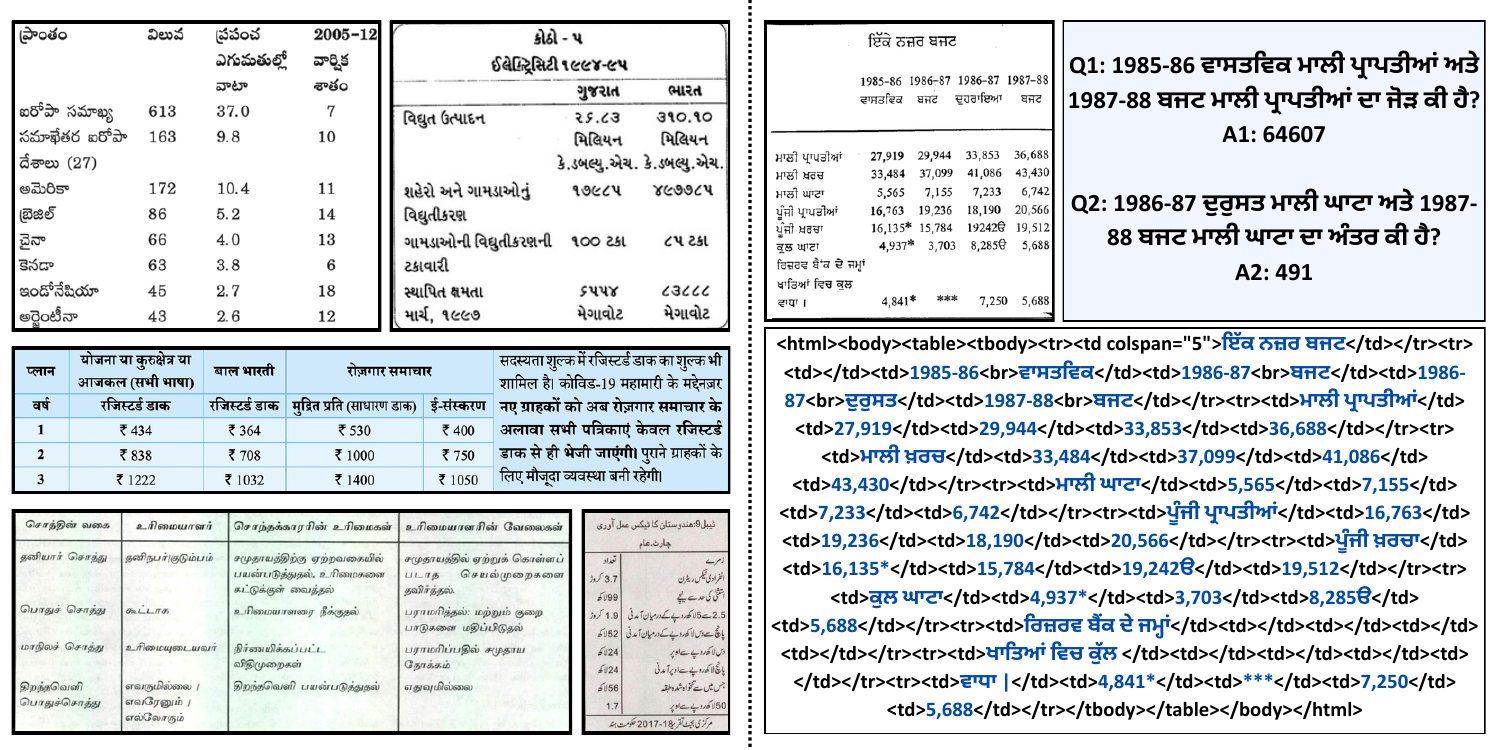}
        \caption{Sample of multilingual tables from \dataset{} dataset showcasing the linguistic and structural diversity across various non-Latin languages.}
        \label{fig:dataset_samples}
    \end{subfigure}

    \vspace{2mm}

    \begin{subfigure}{\linewidth}
        \centering
        \includegraphics[width=0.85\linewidth,trim={12.8cm 0 0 0},clip]{figures/samples.pdf}
        \caption{Punjabi table (up left) from \TRCQA{} with its  HTML representation (down) and two complex reasoning-based question-answer pairs (up right).}
        \label{fig:trcqa_samples}
    \end{subfigure}

    \caption{Representative samples from \dataset{} and \TRCQA{}.}
    \label{fig:benchmarks}
\end{figure}

We introduce \TRCQA{}, a multilingual benchmark for end-to-end TR and CTR over non-Latin tables. It provides 364 tables with reconstructed HTML annotations (including grounded cell bounding boxes) for TR evaluation, together with 714 manually verified reasoning QA pairs spanning Hindi, Telugu, and Punjabi. The benchmark comprises 71 scanned and 293 digitally-born tables collected from government circulars~\cite{mphc, rajbhasha} and multilingual tables in \dataset, capturing diverse layouts, document qualities, and structural complexities encountered in real-world settings. To move beyond simple information extraction, the QA pairs were generated using GPT-oss-20B~\cite{agarwal2025gpt} and manually verified, requiring structural understanding, multi-hop reasoning, aggregation, comparison, arithmetic operations, and logical inference over reconstructed tables. Dataset statistics are summarized in Table~\ref{\TRCQA{} stats}, while representative examples are shown in Figure ~\ref{fig:trcqa_samples}. By jointly supporting multilingual TR and CTR, \TRCQA{} fills a critical gap in multilingual table understanding beyond existing English-centric benchmarks.

\begin{table}[!htb]
\centering
\begin{tabular}{ccc}
\hline
\textbf{Language} & \textbf{No. of Images}  & \textbf{No. of QA Pairs} \\
\hline
Hindi   & 163 & 312 \\
Telugu  & 102 &  204 \\
Punjabi & 99 &  198 \\
\hline
\textbf{Total} & \textbf{364} &\textbf{714} \\
\hline
\end{tabular}
\caption{\TRCQA{} Statistics: the number of table images and corresponding QA pairs across Hindi, Telugu, and Punjabi languages.}
\label{\TRCQA{} stats}
\end{table}

\section{Experiments}
In this section, we describe the masked dataset, followed by the fine-tuning strategy employed for \name{}.

\subsection{Partial Text-Masked Physical TSR Dataset}
\label{masked-dataset}
\begin{figure}[!htb]
    \centering
\includegraphics[width=0.85\linewidth]{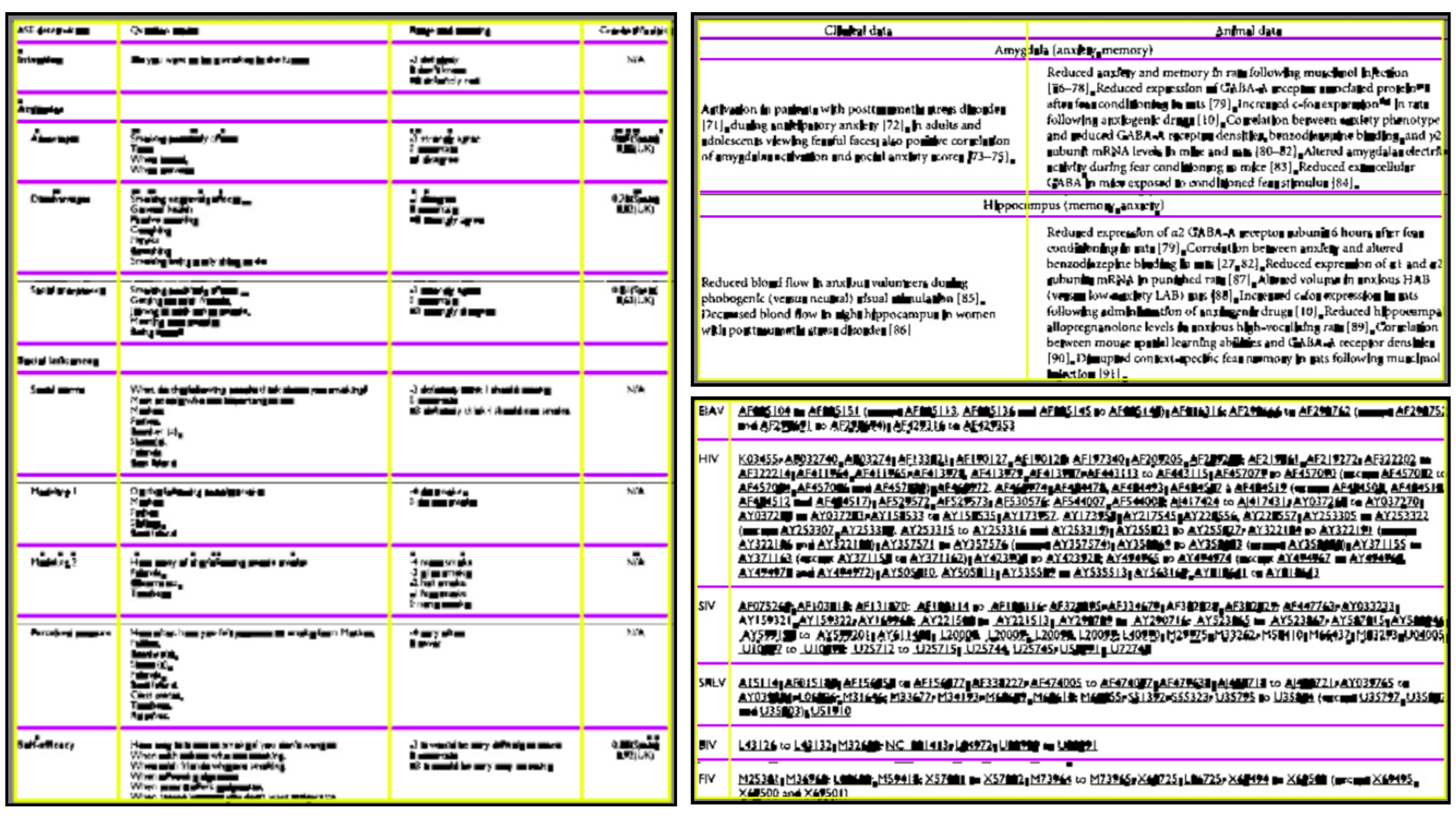}
    \caption{Masked training samples annotated with row boxes (yellow) and column boxes (pink) used for fine-tuning \name.}
    \label{fig:samples}
\end{figure}

To train \name{}, we construct a partially text-masked Physical TSR dataset by transforming existing English table datasets while preserving their row and column annotations. This enables script-agnostic learning without requiring additional multilingual annotations. Our training corpus comprises approximately 1,000 samples from PubTables-1M~\cite{tatr-pub-1m}, 258 samples from ICDAR 2013~\cite{icdar2013-aligned}, and 1,000 contour-masked variants generated from PubTables-1M. Following Algorithm~\ref{alg:masking}, contour-based masking suppresses textual regions while preserving the underlying table geometry, including ruling lines, alignment, and spacing, thereby encouraging the model to rely on structural rather than script-specific cues. To further improve robustness, we apply random grayscale conversion (10\%), Gaussian blur (up to 1 px), and impulse noise (up to 1.25\% pixels) on a few images, resulting in 5,244 training and 498 validation images. A few masked samples are shown in Figure~\ref{fig:samples}.

\begin{algorithm}[!htb]
\caption{Table Image Masking Algorithm}
\small
\begin{algorithmic}[1]
\Require \textit{image}, \textit{threshold} $= 0.01$
\Ensure \textit{masked\_image}

\State $\textit{gray\_image} \gets \textsc{ConvertToGrayscale}(\textit{image})$
\State $\textit{binary\_image} \gets \textsc{OtsuThresholding}(\textit{gray\_image})$
\State $\textit{inverted\_binary} \gets \textsc{InvertImage}(\textit{binary\_image})$
\State $(\textit{width}, \textit{height}) \gets \textsc{ImageDimensions}(\textit{image})$
\State $\textit{contours} \gets \textsc{FindContours}(\textit{inverted\_binary})$
\State $\textit{masked\_image} \gets \textit{binary\_image}$

\ForAll{$\textit{contour} \in \textit{contours}$}
    \State $(x, y, w, h) \gets \textsc{GetBoundingRect}(\textit{contour})$
    \If{$h < \textit{threshold} \times \textit{height}$ \textbf{and} $w < \textit{threshold} \times \textit{width}$}
        \State \textsc{DrawRectangle}$(\textit{masked\_image}, x, y, w, h)$
    \EndIf
\EndFor

\State \Return $\textit{masked\_image}$
\end{algorithmic}
\label{alg:masking}
\end{algorithm}

\subsection{Finalizing \textbf{\name{}} Backbone}
Having established the partially masked Physical TSR dataset, we next identify the most suitable backbone for \name{}. We compare three representative model families: (i) conventional object detectors (YOLOv11), (ii) transformer-based detectors (DETR), and (iii) vision-language models (VLMs), represented by Qwen-2.5-VL (Qwen)~\cite{qwen}, where row and column bounding boxes are generated via prompting. YOLOv11 and DETR are trained on the dataset described in Section~\ref{masked-dataset} using the default training configurations of YOLOv11\footnote{ \url{https://docs.ultralytics.com/models/yolo11/}} and RF-DETR\footnote{ \url{https://rfdetr.roboflow.com/reference/medium/}}, respectively, while Qwen is evaluated using prompt-based bounding-box prediction. Performance is measured on a held-out disjoint set of 100 Hindi tables, providing a direct evaluation of cross-script generalization. Unless stated otherwise, all detectors use a confidence threshold of 0.7 and NMS with an IoU threshold of 0.1. As reported in Table~\ref{tab:backbone}, conventional detectors achieve reasonable localization but struggle to consistently recover row boundaries. Transformer-based DETR substantially improves localization accuracy and structural consistency, particularly for complex layouts. In contrast, prompt-based VLMs perform significantly worse, even when provided with the exact number of rows and columns predicted by SPRINT, highlighting that general multimodal reasoning is insufficient for precise geometric structure estimation. Moreover, VLMs incur considerably higher inference costs and scale poorly across multilingual layouts. Based on these observations, we adopt DETR as the backbone of \name{}, offering the best trade-off between localization accuracy, structural consistency, and cross-script generalization. With both the training data and backbone finalized, we proceed to fine-tune \name{} as described in Section~\ref{sec:ftn}.

\begin{table}[t]
\centering
\small
\setlength{\tabcolsep}{3.3pt}
\renewcommand{\arraystretch}{1.2}
\begin{adjustbox}{width=0.48\textwidth}
\begin{tabular}{|c|ccc|ccc|c|}
\hline
\multirow{2}{*}{\textbf{Model}} &
\multicolumn{3}{c|}{\textbf{Row}} &
\multicolumn{3}{c|}{\textbf{Column}} &
\multirow{2}{*}{\textbf{mAP50}} \\ \cline{2-7}
&
\textbf{AP50} & \textbf{AP} & \textbf{AR} &
\textbf{AP50} & \textbf{AP} & \textbf{AR} & \\
\hline

YOLOv11 &
76.4 & 38.0 & 57.1 &
80.2 & 50.2 & 67.8 &
79.2 \\

DETR &
80.2 & 40.2 & 56.4 &
\textbf{91.4} & 57.2 & 66.2 &
\textbf{85.8} \\
\hline

Qwen &
12.4 & 2.9 & 9.2 &
18.1 & 4.9 & 11.6 &
15.2 \\

Qwen + SPRINT &
10.9 & 2.8 & 9.3 &
17.7 & 4.6 & 11.7 &
14.3 \\
\hline
\end{tabular}
\end{adjustbox}
\caption{Physical TSR performance on Hindi tables for row and column detection across different backbones.}
\label{tab:backbone}
\end{table}

\subsection{Fine-tuning for \textbf{\name}}
\label{sec:ftn}
Having identified DETR as the strongest backbone, we next investigate whether script-agnostic physical TSR can be achieved without any multilingual supervision. We compare three progressively stronger DETR-based variants on the same held-out set of 100 Hindi tables (Table~\ref{tab:model_comparison}): vanilla DETR, TATR (a DETR backbone pre-trained on PubTables-1M for TSR) with row and column classes only, and \name{}, which retains the same DETR backbone as TATR but is further fine-tuned on the partially masked English dataset described in Section~\ref{masked-dataset}. Thus, \name{} is essentially TATR adapted through masking-based fine-tuning, requiring no additional multilingual annotations. The intuition is simple: masking suppresses script-specific textual appearance while preserving table geometry, compelling the model to rely on structural cues such as ruling lines, spacing, and layout instead of language-dependent features. Consequently, although fine-tuned exclusively on augmented English tables, \name{} learns geometry-centric representations that transfer effectively to unseen scripts. We follow the default TATR fine-tuning configuration~\cite{tatr-pub-1m}, resizing input images to $576\times576$ and training for 64 epochs using AdamW ($10^{-4}$ learning rate), linear warm-up, batch size 8, and early stopping. As shown in Table~\ref{tab:model_comparison}, introducing masking-based fine-tuning improves overall cross-script performance, validating our hypothesis that geometry-centric supervision enables robust script-agnostic physical TSR.

\begin{table}[!htb]
\centering
\setlength{\tabcolsep}{4pt}
\renewcommand{\arraystretch}{1.2}
\begin{adjustbox}{width=0.48\textwidth}
\begin{tabular}{|c|ccc|ccc|c|}
\hline
\multirow{2}{*}{\textbf{Model}} & \multicolumn{3}{c|}{\textbf{Row}} & \multicolumn{3}{c|}{\textbf{Column}} & \multirow{2}{*}{\textbf{mAP50}} \\ \cline{2-7}
 & \multicolumn{1}{c}{\textbf{AP50}} & \multicolumn{1}{c}{\textbf{AP}} & \textbf{AR} & \multicolumn{1}{c}{\textbf{AP50}} & \multicolumn{1}{c}{\textbf{AP}} & \textbf{AR} &  \\ \hline
DETR & \multicolumn{1}{c}{80.2} & \multicolumn{1}{c}{40.2} & 56.4 & \multicolumn{1}{c}{\textbf{91.4}} & \multicolumn{1}{c}{57.2} & 66.2 & 85.8 \\
TATR & \multicolumn{1}{c}{82.1} & \multicolumn{1}{c}{69.2} & 77.3 & \multicolumn{1}{c}{85.7} & \multicolumn{1}{c}{\textbf{71.3}} & \textbf{83.9} & 83.9 \\ \hline
\textbf{SPECTRUM} & \multicolumn{1}{c}{\textbf{87.9}} & \multicolumn{1}{c}{\textbf{69.7}} & \textbf{79.9} & \multicolumn{1}{c}{{87.5}} & \multicolumn{1}{c}{58.1} & 71.0 & \textbf{87.7} \\ \hline
\end{tabular}
\end{adjustbox}
\caption{
Comparison of DETR-based Physical TSR models on 100 Hindi tables; \name{} achieves the best performance.}
\label{tab:model_comparison}
\end{table}

\section{Results and Discussion}
\label{sec:results}
We evaluate \name{} on physical TSR, \TR{} on TSR, and eventually on downstream tasks to assess its effectiveness in both English-centric and script-agnostic settings.

\subsection{Performance on Physical TSR}
\begin{figure}[!htb]
    \centering
\includegraphics[width=0.85\linewidth]{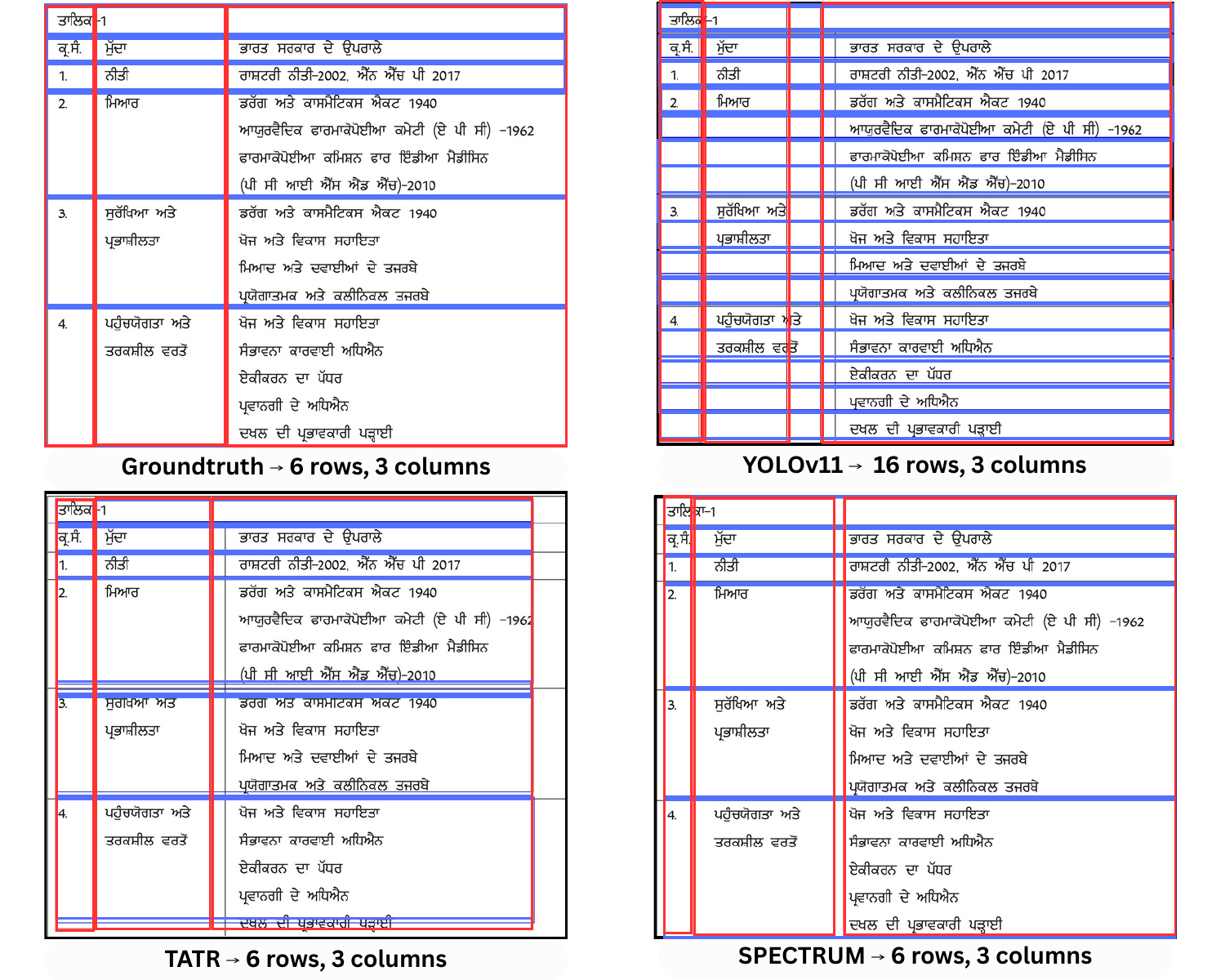}
    \caption{Qualitative physical TSR comparison on a Punjabi table from \dataset{}, showing best localization by \name{}.}
    \label{fig:qual}
\end{figure}

Table~\ref{tab:custard_objdet_metrics} reports Physical TSR performance on \dataset{} using standard object detection metrics. As discussed in the previous section, \name{} consistently outperforms all baselines, improving mAP50 by 2.1 p.p. over TATR while achieving the best overall row detection performance. Although YOLOv11 achieves slightly higher column scores, this largely reflects the IoU-based evaluation. It tends to predict narrower column boxes that overlap well with the annotations, whereas \name{} recovers full-height column extents that better represent the underlying table geometry. As both approaches ultimately assign the same cells, this discrepancy has little effect on downstream tasks. A qualitative Punjabi example from \dataset{} is shown in Figure~\ref{fig:qual}.

\begin{table}[!htb]
\centering
\setlength{\tabcolsep}{3pt}
\renewcommand{\arraystretch}{1.2}
\begin{adjustbox}{width=0.48\textwidth}
\begin{tabular}{|c|ccc|ccc|c|}
\hline
\multirow{2}{*}{\textbf{Model}} &
\multicolumn{3}{c|}{\textbf{Row}} &
\multicolumn{3}{c|}{\textbf{Column}} &
\multirow{2}{*}{\textbf{mAP50}} \\ \cline{2-7}

& \textbf{AP50} & \textbf{AP} & \textbf{AR}
& \textbf{AP50} & \textbf{AP} & \textbf{AR} & \\

\hline



YOLOv11 \cite{yolov11} & 73.1 & 39.4 & 58.4 & 86.7 & \textbf{62.0} & \textbf{76.9} & 79.9 \\
DETR \cite{detr} & 80.9 & 47.3 & 61.9 & 80.7 & 54.3 & 64.3 & 80.8 \\
TATR \cite{tatr-pub-1m} & 80.1 & 53.6 & 65.6 & 87.1 & 56.0 & 70.0 & 83.6 \\

\hline

\textbf{\name} & \textbf{83.0} & \textbf{54.0} & \textbf{67.8} &
\textbf{88.5} & 59.4 & 72.7 & \textbf{85.7} \\

\hline
\end{tabular}
\end{adjustbox}
\caption{Physical TSR performance comparison for row and column bounding box detection on all tables from \dataset.}
\label{tab:custard_objdet_metrics}
\end{table}

\subsection{Performance on TSR}
For multilingual TSR, we employ \TR{} to generate bounding-box-attributed HTML representations of tables. The predicted structure is evaluated using TEDS-S, which measures structural fidelity independent of cell content. Table~\ref{tab:tedss} shows that \TR{} achieves performance comparable to strong end-to-end VLM and sequential TSR pipelines on standard English benchmarks despite being designed primarily for multilingual generalization. \TR{} achieves performance nearly identical to DELTA on standard English benchmarks, as both leverage a similar decoupled TSR pipeline. More importantly, the multilingual results in Table~\ref{tab:custard_tsr_results} demonstrate the effectiveness of our approach: \TR{} consistently outperforms both MTL-TabNet and DELTA across 12 non-Latin languages, yielding an average improvement of 4.9 p.p. in TEDS-S. This confirms that replacing the physical TSR module with our script-agnostic \name{} does not compromise English performance, while yielding substantial gains on multilingual tables.

\begin{table}[!htb]
\centering
\setlength{\tabcolsep}{4pt}
\renewcommand{\arraystretch}{1.05}
\begin{tabular}{|cc|ccc|}
\hline
\multirow{1}{*}{\textbf{Paradigm}} & \multirow{1}{*}{\textbf{Baseline}} &
\textbf{FTN} & \textbf{PTN} & \textbf{P1M} \\ \hline

\multirow{4}{*}{\begin{tabular}[c]{@{}c@{}}End-to-end\\VLM\end{tabular}}
& SmolVLM \cite{smolvlm} & -- & -- & -- \\
& MTL-TabNet \cite{mtl-tab-net} & \textbf{98.8} & \textbf{97.9} & \textbf{96.7} \\
& SmolDocling \cite{nassar2025smoldocling} & 81.0 & -- & 65.0 \\
& Granite Vision \cite{granitevision} & -- & -- & -- \\ \hline

\multirow{2}{*}{\begin{tabular}[c]{@{}c@{}}Sequential\\TSR\end{tabular}}
& EDD \cite{teds} & 90.6 & 89.9 & -- \\
& LGPMA \cite{lgpma} & -- & 96.7 & -- \\ \hline

\multirow{2}{*}{\begin{tabular}[c]{@{}c@{}}Decoupled\\TSR\end{tabular}}
& DELTA \cite{rajput2026delta} & 93.6 & 93.6 & 86.5 \\
& \textbf{\TR{} (Ours)} & 93.8 & 93.8 & 86.5 \\ \hline

\end{tabular}
\caption{TEDS-S comparison across FinTabNet (FTN), PubTabNet (PTN), and PubTables-1M (P1M), showing that \TR{} remains at par with state-of-the-art for TSR on English tables.}
\label{tab:tedss}
\end{table}

\begin{table}[!htb]
\centering
\small
\setlength{\tabcolsep}{6pt}
\renewcommand{\arraystretch}{1.05}
\begin{tabular}{|c|c|c|c|}
\hline
\textbf{Language} & \textbf{MTL-TabNet} & \textbf{DELTA} & \textbf{\TR} \\ \hline
Assamese & 76.5 & 80.3 & \textbf{87.2} \\
Bengali & 61.4 & 68.6 & \textbf{76.7} \\
Chinese & 86.2 & 93.7 & \textbf{94.1} \\
Gujarati & 79.6 & 71.4 & \textbf{80.1} \\
Hindi & 75.1 & 80.9 & \textbf{87.5} \\
Kannada & 67.2 & 73.3 & \textbf{78.7} \\
Malayalam & 81.1 & 86.4 & \textbf{87.6} \\
Oriya & 82.8 & 88.8 & \textbf{91.6} \\
Punjabi & 51.5 & 77.9 & \textbf{82.8} \\
Tamil & 77.8 & 88.9 & \textbf{90.7} \\
Telugu & 82.2 & 84.2 & \textbf{89.6} \\
Urdu & 70.1 & 73.1 & \textbf{79.5} \\ \hline
\textbf{Overall} & 74.3 & 80.6 & \textbf{85.5} \\ \hline
\end{tabular}
\caption{TEDS-S scores for TSR performance on \dataset{} using \TR{} showcasing the improvement in all 12 languages.}
\label{tab:custard_tsr_results}
\end{table}

\subsection{Performance on TR}
To evaluate end-to-end TR, we perform cell-wise OCR on the bounding-box-attributed HTML produced by \TR{}, populating each recovered cell with its recognized content. The reconstructed HTML is then evaluated using TEDS, which jointly measures structural and textual correctness. Table~\ref{tab:teds} shows that leveraging EasyOCR \cite{easyocr}, \TR{} achieves the best performance on FinTabNet among other approaches while remaining on par with other datasets. MTL-TabNet is specifically optimized for English TR and therefore achieves excellent performance on standard English benchmarks. However, the multilingual results in Table~\ref{tab:custard_tr_results} clearly highlight the strength of our design: \TR{} preserves strong English performance while consistently surpassing MTL-TabNet on multilingual TSR, achieving an average improvement of 3.5 p.p. across Hindi, Telugu, and Punjabi tables. TR performance improves when using EasyOCR \cite{easyocr} instead of Tesseract \cite{tesseract}, highlighting the impact of OCR quality on reconstruction. Overall, these results demonstrate that the improvements in multilingual TSR delivered by \TR{} translate directly into more accurate end-to-end table reconstruction.

\begin{table}[!htb]
\centering
\setlength{\tabcolsep}{4pt}
\renewcommand{\arraystretch}{1.05}
\begin{tabular}{|cc|ccc|}
\hline
\multirow{1}{*}{\textbf{Paradigm}} & \multirow{1}{*}{\textbf{Baseline}} &
\textbf{FTN} & \textbf{PTN} & \textbf{P1M} \\ \hline

\multirow{4}{*}{\begin{tabular}[c]{@{}c@{}}End-to-end\\VLM\end{tabular}}
& SmolVLM \cite{smolvlm} & 18.0 & -- & 32.0 \\
& MTL-TabNet \cite{mtl-tab-net} & -- & \textbf{96.7} & \textbf{96.2} \\
& SmolDocling \cite{nassar2025smoldocling} & 52.0 & -- & 88.0 \\
& Granite Vision \cite{granitevision} & 54.0 & -- & 70.0 \\ \hline

\multirow{2}{*}{\begin{tabular}[c]{@{}c@{}}Sequential\\TSR+OCR\end{tabular}}
& EDD \cite{teds} & -- & 88.3 & -- \\
& LGPMA \cite{lgpma} & -- & 94.6 & -- \\ \hline

\multirow{2}{*}{\begin{tabular}[c]{@{}c@{}}Decoupled\\TSR+OCR\end{tabular}}
& DELTA \cite{rajput2026delta} & 55.9 & 54.4 & 54.8 \\
& \textbf{\TR{} + OCR} & \textbf{60.1} & 55.8 & 58.6 \\ \hline

\end{tabular}
\caption{TEDS comparison across FinTabNet (FTN), PubTabNet (PTN), and PubTables-1M (P1M). We use EasyOCR with \TR{}}
\label{tab:teds}
\end{table}

\begin{table}[!htb]
\centering
\small
\setlength{\tabcolsep}{3.3pt}
\renewcommand{\arraystretch}{1.05}
\begin{tabular}{|c|c|ccc|}
\hline
\textbf{TSR} & \textbf{OCR} & \textbf{Hindi} & \textbf{Telugu} & \textbf{Punjabi} \\ \hline

\multicolumn{2}{|c|}{MTL-TabNet \cite{mtl-tab-net}} & 44.3 & 50.1 & 27.4 \\ \hline

\multirow{2}{*}{DELTA \cite{rajput2026delta}}
& Tesseract & 55.9 & 63.2 & 60.5 \\ \cline{2-5}
& EasyOCR & 67.7 & 71.0 & -- \\ \hline

\multirow{2}{*}{\textbf{\TR}}
& Tesseract & 56.0 & 66.1 & \textbf{60.9} \\ \cline{2-5}
& EasyOCR & \textbf{73.8} & \textbf{74.9} & -- \\ \hline

\end{tabular}
\caption{Comparative overview of TEDS on Hindi, Telugu, and Punjabi tables from \TRCQA{} using Tesseract and EasyOCR.}
\label{tab:custard_tr_results}
\end{table}

\subsection{Performance on CTR}
Table~\ref{tab:tablet_results} reports CTR performance on \TRCQA{}, where all VLMs and LLMs are evaluated using identical prompts for a fair comparison. End-to-end VLMs struggle on multilingual CTR, with Qwen3-VL achieving the strongest score (46.8 ANLS), highlighting the difficulty of reasoning directly from multilingual table images. In contrast, \TR{} adopts a divide-and-conquer paradigm by first reconstructing a table and then performing reasoning over it. This decoupling consistently improves performance across all LLMs, demonstrating the value of accurate intermediate table representations for multilingual CTR. Among the \TR{} + LLM baselines, Qwen-3-8B \cite{qwen3_2025} achieves the best performance with 59.7 ANLS, establishing a new state of the art on \TRCQA{}. Beyond improved accuracy, the explicit reconstruction produced by \TR{} makes the reasoning process more interpretable and easier to debug than end-to-end VLMs, where perception and reasoning are tightly coupled. Finally, using the best-performing Qwen-3-8B with ground-truth reconstructed tables yields 84.8 ANLS, suggesting that further advances in multilingual TR can directly translate into stronger downstream reasoning.

\begin{table}[htb]
\centering
\begin{adjustbox}{width=0.48\textwidth}
\begin{tabular}{|ll|c|}
\hline
\textbf{Approach} &
\textbf{VLM/LLM} &
\textbf{ANLS} \\
\hline

\multirow{6}{*}{\textbf{End-to-end VLMs}}
& Molmo-7B-D-Instruct~\cite{deitke2024molmopixmoopenweights} & 19.0 \\
& Phi-4 Multimodal~\cite{microsoft2025phi4minitechnicalreportcompact} & 21.2 \\
& Llama 3.2 Vision 11B~\cite{llama32_2024} & 27.7 \\
& GLM-4.1V-9B-Thinking~\cite{vteam2026glm45vglm41vthinkingversatilemultimodal} & 42.1 \\
& Gemma 4 Vision~\cite{gemmateam2026gemma4technicalreport} & 42.1 \\
& Qwen3-VL-8B-Instruct~\cite{qwen3vl_2025} & 46.8 \\

\hline

\multirow{6}{*}{\textbf{\TR{}} + \textbf{LLM}}
& DeepSeek-R1~\cite{deepseekr1_2025} & 46.5 \\
& Llama-3.2-11B-Vision~\cite{llama32_2024} & 51.4 \\
& Llama 3.3 8B~\cite{llama33_2024} & 51.5 \\
& Qwen-3.5-9B~\cite{qwen35blog} & 52.0 \\
& Gemma-4-12B~\cite{gemmateam2026gemma4} & 56.9 \\

& Qwen-3-8B~\cite{qwen3_2025} & \textbf{59.7} \\

\hline

\multirow{1}{*}{\textbf{Skyline (Using GT)}}
& Qwen-3-8B~\cite{qwen3_2025} & 84.8 \\

\hline
\end{tabular}
\end{adjustbox}
\caption{CTR performance comparison on \TRCQA{} using ANLS.}
\label{tab:tablet_results}
\end{table}

\section{Conclusion}
We introduce \name{}, a script-agnostic physical TSR model that localizes rows and columns as bounding boxes. Through selective masking-based augmentation, it suppresses script-dependent cues and learns representations that transfer robustly across scripts. Building on this, we propose \TR{}, a decoupled TSR framework that integrates \name{} with logical TSR to produce complete table structures for downstream applications. We further present \dataset{}, the first benchmark for multilingual TSR across 12 non-Latin languages, along with \TRCQA{}, the first benchmark for multilingual TR and CTR. Experiments show that \TR{} matches state-of-the-art TSR performance on English benchmarks while substantially outperforming prior methods in multilingual settings. Coupled with OCR, the structured outputs of \TR{} further improve TR and CTR, demonstrating that stronger structure modeling directly translates to better downstream understanding. Together, \TR{}, \name{}, \dataset{}, and \TRCQA{} establish a strong foundation for multilingual table understanding and reasoning.
\section{Limitations and Future Work}
Although \name{} and \TR{} substantially advance script-agnostic TSR, end-to-end reconstruction remains limited by OCR quality. By decoupling structure (TEDS-S) from text recognition, we identify multilingual OCR as the primary bottleneck and a key direction for future improvement. Extending language coverage beyond the current three languages is another important next step. Finally, advancing CTR through richer table representations and structure-aware prompting strategies built on top of \TR{} presents a promising research direction.

\section{Acknowledgement}
\label{sec:Acknowledgement}

We gratefully acknowledge the Government of India for its support through MeitY and DST, via their initiative BharatGen, which were instrumental in enabling this work. Jahanvi Rajput’s PhD is supported by the Prime Minister’s Research Fellowship (PMRF).

{
    \small
    \bibliographystyle{ieeenat_fullname}
    \bibliography{main}
}

\end{document}